\documentclass[runningheads]{llncs}
\usepackage[T1]{fontenc}
\usepackage{amsmath} 
\usepackage{graphicx}
\usepackage{acronym}
\usepackage{float}
\begin{document}
\title{Infectious Disease Forecasting in India using LLM's and Deep Learning}
%
%
\author{Chaitya Shah\inst{1,2}\and
Kashish Gandhi\inst{1,3}\and
Javal Shah\inst{1,4}\and
Kreena Shah\inst{1,5}\and
Nilesh Patil\inst{1,6}\and
Kiran Bhowmick\inst{1,7}
}
%
%
\institute{Department of Computer Engineering, Dwarkadas J. Sanghvi College of Engineering, Mumbai, India\\ \and
chaitya0623@gmail.com\\ \and
kashishgandhi6112003@gmail.com\\ \and
javalshah2003@gmail.com\\ \and
kreenashah61260@gmail.com\\ \and
nilesh.p@djsce.ac.in\\ \and
kiran.bhowmick@djsce.ac.in
}
\maketitle              
\begin{abstract}
Many uncontrollable disease outbreaks of the past exposed several vulnerabilities in the healthcare systems worldwide. While advancements in technology assisted in the rapid creation of the vaccinations, there needs to be a pressing focus on the prevention and prediction of such massive outbreaks. Early detection and intervention of an outbreak can drastically reduce its impact on public health while also making the healthcare system more resilient. The complexity of disease transmission dynamics, influence of various directly and indirectly related factors and limitations of traditional approaches are the main bottlenecks in taking preventive actions. Specifically, this paper implements deep learning algorithms and LLM's to predict the severity of infectious disease outbreaks. Utilizing the historic data of several diseases that have spread in India and the climatic data spanning the past decade, the insights from our research aim to assist in creating a robust predictive system for any outbreaks in the future.

\end{abstract}

\section{Introduction}
After witnessing many epidemics over the course of the century, our healthcare systems got a massive blow after its failure to predict and suppress the Corona Virus pandemic of 2019. This unprecedented event not only exposed the shortcomings of the system but also underscored the need for an accurate and reliable predictive system for the occurrence of outbreaks. The importance of predictive modeling in preparing for such epidemics cannot be overstated. 

Predictive modeling plays a vital role in preparing and protecting the masses from such chaotic and harmful events, be it natural disasters or disease outbreaks. Data has been one of the most valuable resources that we have gathered in this technological era however, without leveraging it, it would be of no use or benefit. By analyzing historic data of the spread of diseases, their transmission and related factors like climatic data over a specific time period, researchers can develop sophisticated models to forecast the spread of infectious diseases.

In this context, our research utilizes historical data about the spread of various diseases and the climatic conditions during that period, focused on India specifically, to develop a predictive model for future outbreaks. This model will pave the way for many more proactive and efficient approaches to control the spread of diseases by contributing to a larger objective of strengthening disease control methods.

\section{Literature Review}
There have been consistent research in the field of healthcare and prediction of diseases using various machine learning models. This paper aims to enhance the models and methodologies used by some of those studies and develop a new more efficient predictive model.

Sina F. Ardabili at el. [1] demonstrates the development of a model to predict Covid-19 outbreaks using ML. Using evolutionary algorithms like GA, PSO, GWO cost functions are solved to estimate the parameters. GWO results in the higher accuracy and lower processing time compared to the other algorithms. Moreover, it highlights the logistic model to be accurate out of all the different mathematical models used like logistic, linear, logarithmic, power, cubic, exponential. Understanding the complexities between outbreaks in different countries, the study explains that global models with generalization abilities would not be feasible and deep studies are required in order to improve the efficiency of the current prediction model. Md. Ehtisham Farooqui at el. [2] presents the development of a diease prediction system using SVM and MLR. The system aims to predict the diseases based upon symptoms, reducing the time taken for diagnosis and providing appropriate  treatment. It emphasizes on the potential of SVM and MLR to accurately predict diseases based on symptoms and underscore the need for further research in the predictive modeling field to enhance the effectiveness of healthcare. Furthermore, there is thorough emphasis on utilizing structured datasets to train the disease predicting systems in order to achieve higher accuracy in disease prediction.

Rinkal Keniya at el. [3] proposes a disease prediction system using ML algorithms to accurately diagnose diseases. It draws these predictions based upon the gender, age and symptoms. Multiple ML algorithms are utilized and KNN algorithm demonstrated the highest accuracy at 93.5\%. Decision trees, KNN, Naive Bayes algorithm and RUSBoosted trees are used to process the dataset and predict diseases. The study draws comparisons with existing research papers, portraying the superior performance of the suggested approach. Min Chen at el. [4] presents a method for disease prediction using machine learning over big data from healthcare communities. The paper discusses a critical issue of chronic diseases and its economic impact on the healthcare systems. Considering the complexities of disease patterns in various regions, it highlights the need to use big data for disease prediction and risk management. This paper proposes a CNN based Multimodal Disease Risk Prediction algorithm which combines structured and unstructured data, refractors missing data using latent factor modeling and using CNN selects relevant features.

Durga Mahesh Matta at el. [5] demonstrates the prediction of Covid-19 using machine learning techniques. The study compares three different algorithms namely RF, ANN and SVM and shows RF to be the most accurate out of all. Moreover, the study explains the importance of selecting the right features and its direct impact on the results obtained while disregarding the features that do not affect the results, in order to simplify the process. Godson Kalipe at el. [6] proposes the prediction of malarial outbreak using machine learning and DL approach. The paper predicts the malarial outbreak in Andhra Pradesh, India by analyzing atmospheric factors and climatic data. The data is trained on various classification algorithms namely KNN, Random Forest, SVM, XGBoost, Logistic Regression, Neural Network and Naive Bayes. Using precision, recall and accuracy as the comparison factors, XGBoost clearly outperforms other algorithms with 96.26\% accuracy and 93.89\% recall as shown in this study. However, precision being 91.82\% leaves way for more thorough research in this domain to increase the effectiveness of the prediction system.

Juhyeon Kim at el. [7] demonstrates the development of infectious disease outbreak prediction using media articles with machine learning models. This paper uses mathematical models like the SIR model to simulate the spread of diseases. Moreover, it explores the use of media articles to predict outbreaks and successfully estimates disease dynamics based on temporal topic trends. Sangwon Chae at el. [8] presents predicting infectious diseases using DL and big data. It underscores the problems faced due to conventional reporting systems and emphasizes on the need to utilize big data sources like internet search queries and environmental factors like weather in predictive modeling. It shows the comparison between DL models and proves its strength over linear models, LASSO (Least Absolute Shrinkage and Selection Operator) and ARIMA (AutoRegressive Integrated Moving Average) models in previous studies. Furthermore, it conveys the pressing need to thoroughly test for the models with highest explanatory power.

Muhammad Adnan Khan at el. [9] forecasts the influenza pandemic using machine learning. This paper focuses on the threats due to influenza pandemic particularly the H1N1 pandemic of 2009. It highlights the role of machine learning in being able to accurately predict the influenza pandemics and assist the authorities making precise decisions, provide necessary treatment options and quarantine procedures. Nurul Absar at el. [10] presents the efficacy of DL based LSTM model in forecasting the outbreak of contagious diseases. The paper demonstrates how MinMaxScaler was used to preprocess data and LSTM networks predict the trend of Covid-19 in Bangladesh. This focuses on predictive modeling of Covid-19 outbreaks in places that face resource constraints. 

Santangelo OE at el. [11] discusses various studies on the use of machine learning models to predict infectious diseases such as COVID-19, Zika, hepatitis E, hand, foot, and mouth disease, influenza, and malaria. The studies compare different models and techniques, such as LSTM, ARIMA, SVM, and self-attention, to improve predictive accuracy. The results show that machine learning models can effectively predict infectious disease trends and that certain models, such as LSTM, perform better than others. The document also highlights the importance of space/time resolution, order of magnitude modeled populations, funds, and conflicts of interest in these studies. Scholastica Ijeh at el. [12] comprehensively examines the field of predictive modeling for disease outbreaks. It explores the diverse data sources used, such as epidemiological, environmental, social media, and mobility data, highlighting the importance of integrating multiple sources to improve accuracy. The paper discusses various modeling techniques like statistical models, machine learning, and network analysis, along with their strengths, limitations, and suitability for different diseases and data types. It evaluates methods to assess model accuracy, such as sensitivity and specificity, while discussing challenges in model validation like data quality and dynamic disease spread. Rather than presenting new results, the paper synthesizes existing research to provide a broad perspective on data sources, modeling approaches, challenges, and future prospects for predictive modeling of disease outbreaks.

The above mentioned papers showed the need to improve such predictive models and encouraged us to pursue deeper research on the same lines. Using insights from these different papers, our research draws inspiration to focus on the impact of climatic factors on disease transmission and implements machine learning and DL based approach to create a predictive model. 

\section{Proposed Methodology}

\subsection{Creating the Dataset}\label{AA}
In the meticulous assembly of our dataset, our foremost concern was ensuring the reliability of our information, sourced exclusively from reputable and official websites. We focused on three important areas: first, we gathered detailed information about nine different diseases; second, we analyzed weather data collected using precise web scraping techniques; and third, we delved into the demographics of the people who were infected. This comprehensive approach allows us to understand the intricate connections between disease outbreaks and various influencing factors. By exploring these different aspects, we gain a deeper understanding of the complex dynamics at play in our study.

\paragraph{Disease Dataset}
We had to collect this data from various official sources [13][14][15] as there was no aggregate data available on the internet for all such diseases. Except Covid-19 and Influenza, other data was collected from [13]. It is well preserved and organized, allowing detailed studies of various diseases and therefore nuanced understandings of public health trends. We have covered the entire spectrum from Typhoid with historical cases and not so widespread to the Covid-19 pandemic, so we take a closer look at the Indian healthcare system. During our investigation, we encountered unique insights into symptom-related challenges, particularly during the initial phase of database design. It was essential to adopt an innovative perspective due to the absence of overt symptoms in the provided list. To address this, a sequential reading approach was employed as a foundational step to comprehensively examine and structure the data. The symptoms dataset utilized initially originated from [16], following which it underwent further processing to refine its utility. Initially, we started by creating a dictionary-based format to organize the data into basic structures. However, the qualitative group posed a unique challenge which meant that more sophisticated approaches had to be developed in the earlier steps as far as extraction was concerned. This methodological approach produced coherent and complete data. The preprocessing steps not only addressed the complexity of symptom classification, but also enhanced the analytical utility of the dataset by considering and beyond a variety of other populations this inclusion of personal illness data seamlessly reflected the interrelated health challenges faced by the population.

\begin{figure}[htp]
    \centering
    \includegraphics[width=8.6cm, height=6cm]{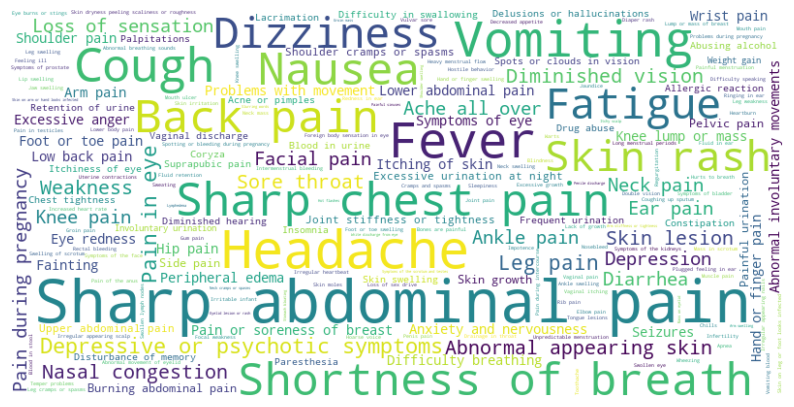}
    \caption{Wordcloud for Most Common Symptoms}
    \label{fig:galaxy}
\end{figure}

Each disease category in the dataset covers important health care outcomes, contributing to nuanced understandings of population health trends. Notably, the Covid-19 dataset [15] covers cases and deaths from 2020 to 2024. The WHO has been releasing updates on confirmed cases and fatalities globally since December 31, 2019. From that date until March 21, 2020, the information was gathered from official communications following the IHR, 2005, supplemented by publications on official health ministry websites and social media platforms. Starting March 22, 2020, the data has been consolidated from WHO region-specific dashboards or directly reported to the WHO. The WHO updates its data on a weekly basis. In influenza, this dataset [14] covers monthly cases in India from 2009 to 2023, providing a detailed breakdown of cases, deaths and disease group. It should be noted that the influenza dataset was used primarily for testing purposes. Annual disclosure of data, which sheds light on the rate of death caused by unsafe sanitation per 100,000 people. This highlights the seriousness of health issues associated with inappropriate sanitation practices and serves as a resource value for policy makers. The database examines the impact of influenza diseases and provides annual data from 2000 to 2019 on mortality in India. This information helps provide insight into the broader public health implications of influenza illnesses. Recorded annual dengue cases from 1990 to 2019 are examined to understand the spread and incidence of this mosquito-borne disease over the years Covering invasive non-typhoid Salmonella outbreaks from 1990 to 2019 adopts a metric of typhoid per 100,000 population , of disease burden Provides a standardized approach. The longitudinal perspective on tuberculosis is gleaned from yearly data spanning 2000 to 2019, providing a comprehensive insight into the tuberculosis situation in India. Historical data on cholera, from 1950 to 2020, documents reported cases, revealing insights into the occurrence and impact of cholera over several decades. Pneumonia, with annual records from 1990 to 2019, focuses on the number of deaths due to pneumonia and lower respiratory diseases, contributing significantly to our understanding of the burden of pneumonia on public health. This list of diseases culminates of 29177 unique rows, an integrated repository for advanced analysis. 

This convention facilitates the comparative study of diseases, and gives us a comprehensive view of the health of India. Notably, all possible diseases are considered, from the much-hyped Covid-19 to the historically significant typhoid. While emphasizing the breadth of the description, it highlights the significant difference in the nature of the spread of Covid-19 compared to other diseases. The influenza dataset played the main role in testing, and the other eight datasets were used for training.

\paragraph{Yearly Weather Dataset}
This dataset was curated by collecting and processing historical weather data from within the Weather.com API [17], zeroing in on the geographic coordinates VIDP:9:IN. Spanning from January 1, 2009, to February 23, 2024, the resultant dataset encompasses a myriad of weather parameters for each day. The standardized dataset contains key characters such as "Date","Average Temperature (C)","Average Temperature (F)","Most Repeated Weather Phrase","Average Wind Speed (mph)","Average Wind Speed (kph)","Average Wind Degree","Most Repeated Wind Direction","Average Pressure","Average Dew Point","Average Heat Index","Average Visibility","Most Repeated Cloud Cover","Average UV Index". Each entry contains a detailed map of the atmosphere on the corresponding day, providing a detailed view 

Our script, which is run by a high-performance system, dynamically parsed the specified date, sorted API requests for each day and extracted many weather observations into complex calculations, providing average temperatures in Celsius and Fahrenheit came in, showed frequent weather terms and calculated average wind speed. In addition to miles per hour and kilometers per hour, the text measured prevailing wind direction, average atmospheric pressure, dewpoint, temperature of characteristics, visibility, and UV characteristics and also assessed primary cloud cover for each recorded day. This dataset, a valuable resource resulting from our proficient data compilation, serves a dual purpose. Beyond providing detailed weather insights, it acts as a foundational element for in-depth analyses and modeling endeavors, specifically geared towards understanding the correlation between weather patterns and disease outbreaks. By integrating weather data with epidemiological records, this dataset facilitates a holistic exploration of trends and patterns, contributing to a nuanced understanding of the interplay between environmental conditions and the dynamics of disease outbreaks over the specified temporal horizon.

\begin{figure}[H]
    \hspace*{-1.5cm} 
    \centering
    \includegraphics[width=14cm]{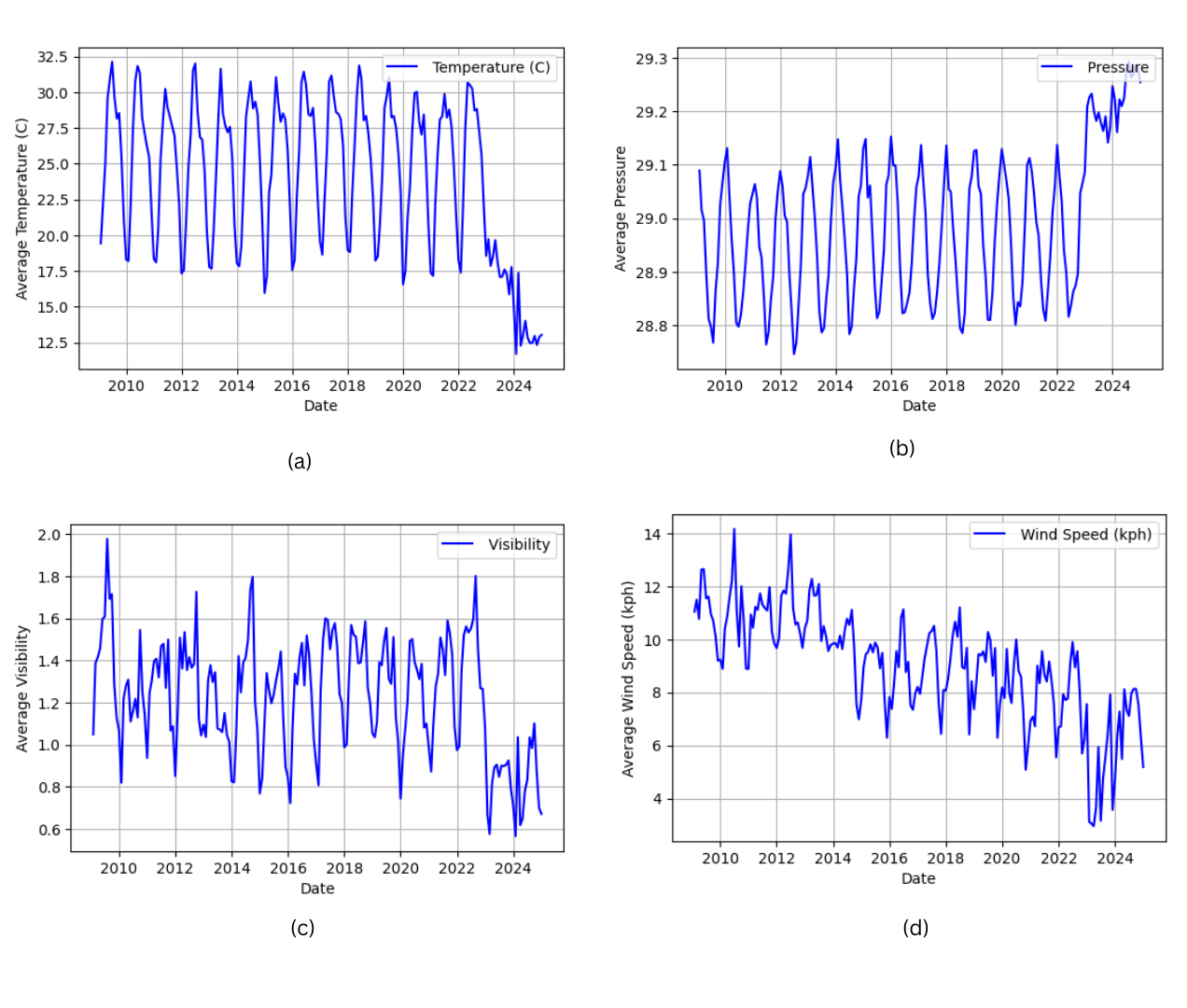}
    \caption{Weather trends for 2(a) Yearly Temperature, Fig 2(b) Yearly Pressure, Fig 2(c) Yearly Visibility, Fig 2(d) Yearly Wind Speed from January 2009 - January 2024.}
    \label{fig:galaxy}
\end{figure}

\paragraph{Demographics Dataset}
We adeptly merged a demographics dataset, expanding the spectrum of insights beyond symptomatology. We subjected the dataset to meticulous processing, augmenting its analytical prowess. Notably, data acquisition for demographics was also facilitated through [16]. The initial dataset, encapsulated crucial disease-related details such as codes, names, symptoms, descriptions, test procedures, and additional data. A Python script adeptly navigated each line, systematically structuring the data into a DataFrame. Subsequent preprocessing steps meticulously refined and organized the information for optimal clarity.

Symptoms, descriptions, test procedures, and additional data underwent precise extraction and structuring, employing string manipulation techniques and regular expressions for uniformity. This process ensured a coherent dataset, facilitating interpretability and seamless integration of demographic attributes. The demographics dataset seamlessly merged with the symptoms dataset, enriching it with attributes like disease descriptions, test procedures, medication details, medications, and risk factors.

The final columns of this integrated dataset include: code, name, symptoms, description, test\_procedure, medication\_desc, medications, symptom\_desc, \newline risk\_years, less\_risk\_years, high\_risk\_race\_ethnicity, high\_risk\_gender,\newline less\_risk\_race\_ethnicity, and less\_risk\_gender. A comprehensive analysis of demographic aspects ensued, visualizing age distribution, gender representation, and race/ethnicity demographics through insightful histograms and bar charts. Additionally, delving into symptom frequencies uncovered the top 10 symptoms, providing valuable insights for a nuanced understanding of disease manifestations. This data integration and analysis endeavor, signifies an approach to extract meaningful insights from intricate datasets. The outcome serves as a foundation for robust epidemiological studies and informed healthcare decision-making processes.

\begin{figure}
    \centering
    \includegraphics[width=\textwidth]{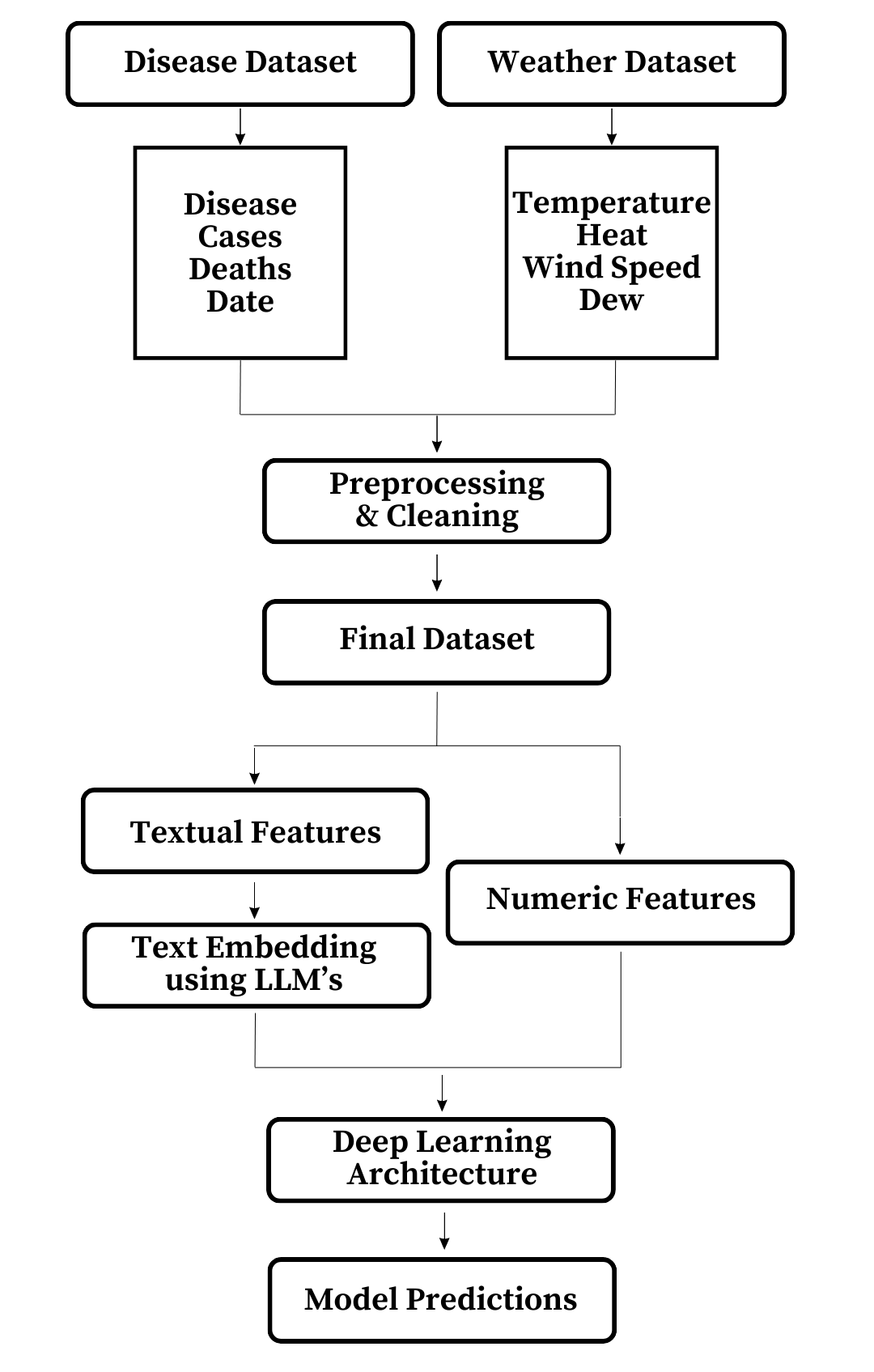}
    \caption{Model Architecture}
    \label{fig:galaxy}
\end{figure}

\subsection{Text Embeddings}
For all the columns in our dataset of the type string (Disease, Symptoms and Most Repeated Weather Phrase), it was important to generate embeddings as the model is not capable of reading text data. Thus, our inital approach was to use One-hot Encoding for the same. Due to the high amount of unique values in the individual columns, it lead to high dimensionality. When it comes to symptoms, having a semantic similarity for similar symptoms is important. But One-hot encoding is incapable of handling semantic relationships and relationships between words. Another main disadvantage is the loss of sequential information. In order to overcome these problems, we thought of using a pretrained LLM, that is capable of inculcating all these important features.

In the realm of evolving technologies in NLP, the need for application oriented as well as effective model is relentless. We have used the pretrained 'distilbert-base-uncased' model for our approach. DistilBERT is a distilled version of the base model of BERT. The model is also uncased, that is, it is not case sensitive. This model is a transformer model which is smaller than the BERT Model, but still faster. It was trained on the same corpus as the BERT base model, in a self-supervised fashion. In order to train this model, masked language modeling is used, which is different from traditional neural networks that sequentially see words one after the other.

Thus, in order to ensure low dimensionality of the dataset, and to preserve the semantic similarity between different symptoms, mentioned separately, we conclude on using this model. Thus, by passing all the symptoms as the input text, high dimensionality vectors (embeddings) are generated, which represents each word, or token with respect to the entire context of the input. These embeddings are further mapped with the training dataset.

\subsection{Deep Learning Model}
Hakizimana Leopord, Dr. Wilson Kipruto Cheruiyot, Dr. Stephen Kimani, [18], in their paper emphacized on the need for developing robust model for predicting disease outbreak in datasets spanning various countries by filling the existing data mining technique gaps where the majority of models are relaying on singular approaches, due to which their accuracies in prediction are not maximized for achieving expected results. This gave us the motivation to create a new model, that not only takes previous data into consideration, but also takes the symptoms and the relevant climatic conditions at those particular timestamps into consideration. Another paper by Sunita Tiwari, Sushil Kumar and Kalpna Guleria [19] predicted the outbreak of coronavirus disease-2019 in India. Their model predicted the potential cases for the next quarter, trying to showcase the overall impact, and to create an alarm amongst citizens to shed light on controlling the situation. 

Visualizing the outputs of the regressive models created in the above mentioned papers, we felt that there is need to utilize a newer form of model. We thought of using DL models for the same. DL models can capture complex non-linear relationships within the data. Infectious diseases often exhibit non-linear patterns influenced by various factors like density, climate and human behaviour. DL models excel at representing these intricate relationships, which might be challenging for classical linear regression models. DL models can automatically learn hierarchical representations of features from the input data. Thus, while regression models typically require manual feature engineering, DL models can directly extract informative features from the raw data, potentially predicting accuracy and generalizing performance. Infectious disease outbreaks usually unfold over time, and in such situations, DL models are well-suited for modeling sequential data and capturing temporal dependencies. Thus, based on historical data, more accurate and timely predictions can be made. As the dataset was collected from various sources, it can contain noise and missing data. DL models can learn meaningful representations from partially observed or noisy data, mitigating the impact of data imperfections on prediction performance. Furthermore, DL architectures can easily be customised according to the problem by fine tuning pre-trained models on domain-specific knowledge. This flexibility allows us to tailor our model into our data potraying the unique characteristics of infectious diseases.

Our initial approach to predicting disease outbreaks was to use the numerical features extracted from the dataset. For this, a DL architecture was implemented. This architecture consisted of 5 dense layers with ReLU Activation Function. In order to prevent overfitting to the training data, regularization techniques such as the L2 regularization were employed.

Our final approach involved integrating both numerical features as well as text embeddings for symptoms. In this hybrid model, both, numerical features and text embeddings were concatenated to create a combined feature vector. This combined vector was then fed into a similar DL architecture as the numeric model. The hybrid model aimed to leaverage both numerical and textual information for improved accuracy.

\vspace{10pt}
\textbf{Concatenate Layer:}
\[
x_{\text{concatenated}} = \text{Concatenate}(x_1, x_2, ..., x_n)
\]

\textbf{Dense Layer (with ReLU activation):}
\[
z_i = \text{ReLU}(w_i \cdot x + b_i)
\]

\textbf{L2 Regularization Term:}
\[
\text{Regularized Loss} = \text{Original Loss} + \lambda \sum_{i,j} W_{i,j}^2
\]

\textbf{Adam optimizer:}
\[
\theta_{t+1} = \theta_t - \frac{{\eta}}{{\sqrt{\hat{v}_t} + \epsilon}} \cdot \hat{m}_t
\]

where:
\begin{align}
\hat{m}_t &= \frac{m_t}{1 - \beta_1^t} \\
\hat{v}_t &= \frac{v_t}{1 - \beta_2^t} \\
m_t &= \beta_1 m_{t-1} + (1 - \beta_1) g_t \\
v_t &= \beta_2 v_{t-1} + (1 - \beta_2) g_t^2
\end{align}

Here, 
\begin{itemize}
\item $\theta_t$ is the parameter vector at iteration $t$,
\item $\eta$ is the learning rate,
\item $\epsilon$ is a small value to prevent division by zero,
\item $\beta_1$ and $\beta_2$ are exponential decay rates for moment estimates,
\item $m_t$ and $v_t$ are first and second moment estimates, respectively,
\item $g_t$ is the gradient at iteration $t$.
\end{itemize}

\section{Results}
The visualizations of climatic factors as seen in Fig. 2 signifies the potential relationship between environmental conditions and disease outbreaks. These trends aided in understanding the influencing factors and developing accurate predictive models. The model is able to effectively capture the complex non-linear relationships between disease outbreaks and various factors like climatic conditions, symptoms, and historical disease data. The prediction results, as depicted in Table I and Fig. 4, demonstrate the model's capability to forecast disease outbreaks with reasonable accuracy. The main objective is to understand the model's ability on real world data. To check if it is suitable, we thought of testing it on an existing disease (Influenza Dataset) for evaluating the model's performance and adding validity to the results, proving it's vitality over future real world data. Influenza was chosen as it is a widespread disease with significant impact, yet not as severe as COVID-19 or as chronic as tuberculosis. The use of standard evaluation metrics like MAE, MSE, RMSE, and R-Squared allows for comparisons with other predictive models and provides a quantitative measure of the model's accuracy. The evaluation metrics, such as a high R-Squared value of 0.95, indicate that the model can capture a significant portion of the variance in the data, validating its predictive performance. Incorporating symptom information, along with numerical data, into the hybrid DL model improves its predictive accuracy.

\begin{table}[htbp]

\begin{center}
\renewcommand{\arraystretch}{1.5}
\begin{tabular}{|c|c|}
\hline
\textbf{Metric} & \textbf{Value}\\
\hline
{Mean Absolute Error} & {1846.39}\\
\hline
{Mean Squared Error} & {13367089.49}\\
\hline
{Root Mean Squared Error} & {3656.10}\\
\hline
{R-Squared} & {0.95}\\
\hline
\end{tabular}
\label{tab1}
\vspace{10pt}
\caption{Result Metrics}
\end{center}
\end{table}

\vspace{-15pt}

The predictive capability of the developed model can assist public health officials and policymakers in proactively preparing for potential disease outbreaks, allocating resources, and implementing preventive measures.
Early detection and intervention based on the model's predictions can help mitigate the impact of outbreaks on public health and reduce socioeconomic implications.

\begin{figure}[htp]
    \centering
    \includegraphics[width=8cm]{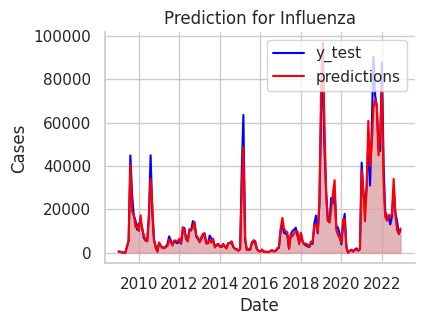}
    \caption{Prediction Disease Outbreak based on Climatic Conditions}
    \label{fig:galaxy}
\end{figure}

\section{Conclusion}
This research has demonstrated the potential of predictive modeling in the field of public health, with crucial focus on infectious disease outbreaks. The 2019 Corona Virus pandemic served as a brutal reminder of the shortcomings in the technology related to the field of medicine and health and put emphasis on the need for proactive measures to mitigate future outbreaks. Using the past data spanning over a decade, related to previous infectious disease outbreaks and weather conditions of India, this study has created a predictive model for anticipating the spread of any future outbreaks. By implementing machine learning algorithms and analytical techniques, we gained valuable insights into the spread of diseases and its associated factors.

The developments in this research have widespread benefits in the field of health and sanitation. With more reliable models to forecast an epidemic, public health officials can take the necessary actions to safeguard the population from fatal outbreaks. Necessary preparations and actions can be taken by the authorities to direct the development of a robust healthcare system in the right direction. Policymakers and government officials can precisely estimate the future requirements for procurement of funds for the same purposes. All of these benefits not only safeguard the well-being of the population but also reduce the socioeconomic implications of epidemics.

This research boosts the progress of the currently undertaken efforts to strengthen the healthcare systems worldwide and underscores the importance of leveraging technology and data driven approaches in combating infectious diseases. Maximizing the use of technology in the sector of public health, we can confidently strive towards a future with more resilient public health and even more suppressed impact of disease outbreaks.

\section{Future Scope}
The following are some findings and avenues for more detailed research that we explored during the development of this predictive model.
\begin{itemize}
\item Integration of Data Sources: The current research has been focused on India. Create a more elaborate model by integrating data from various countries. Include more associated factors and some loosely related factors such as socioeconomic factors, travel patterns, demographic information, etc. 
\item Predicting Demographics: Just like the way outbreak was predicted in this paper, targeted age group, gender can also be predicted with the help of our demographics dataset. Making the prevention systems more cautious. 
\item Real Time Monitoring: Develop real time monitoring systems that use predictive models to detect and forecast outbreaks as they emerge. Incorporate data from social media and surveillance systems to enable timely public health interventions.
\item Decision Support Tools: Develop a user interface that enables stakeholders to utilize the predictions and derive insights from it. Create a user-friendly dashboard to assist in data-driven decision making for all stakeholders.
\item External Validation Studies: Improve credibility of predictive model by partnering with healthcare agencies and performing studies to validate the results of the model. Expand the reach to different demographics and time-periods to assess the scope of the model.
\end{itemize}

\begin{credits}

\subsubsection{\discintname}
The authors declare that they have no conflict of interest and no competing interests. All authors contributed to the study conception and design. Material preparation, data collection and analysis were performed by all authors. This project has no funding. The manuscript was written and revised by all authors on previous versions of the manuscript. All authors read and approved the final manuscript.
\end{credits}
%
%
%
%

\end{document}